\pdfoutput=1 % arXiv: compile with pdflatex
\documentclass[runningheads]{llncs}

\usepackage[T1]{fontenc}

\usepackage{soul}
\setuldepth{foobar}

\usepackage[table]{xcolor} % for \cellcolor
\usepackage{booktabs}      % for nicer tables
\usepackage{pgf}           % for calculations
\usepackage{fontawesome5}
\usepackage{pifont}
\usepackage{svg} %moritz
\usepackage{longtable}
\usepackage{booktabs}
\usepackage{multirow}
\usepackage{adjustbox}
\usepackage{upquote}
\usepackage{tikz}
\usepackage{pgf-pie}

\usepackage{textcomp}
\makeatletter
\renewcommand{\@fnsymbol}[1]{*}  % Change to bold and large asterisk
\makeatother

\makeatletter
\newif\ifcvpr@inappendix
\cvpr@inappendixfalse

\newcommand{\CVPRAppendixTOCMarker}{}

\newcommand{\AppendixOnlyTOC}{%
  \begingroup
    \let\cvpr@orig@section\l@section
    \let\cvpr@orig@subsection\l@subsection
    \let\cvpr@orig@subsubsection\l@subsubsection
    \let\cvpr@orig@paragraph\l@paragraph
    \let\cvpr@orig@subparagraph\l@subparagraph
    \renewcommand{\l@section}[2]{\ifcvpr@inappendix \cvpr@orig@section{##1}{##2}\fi}
    \renewcommand{\l@subsection}[2]{\ifcvpr@inappendix \cvpr@orig@subsection{##1}{##2}\fi}
    \renewcommand{\l@subsubsection}[2]{\ifcvpr@inappendix \cvpr@orig@subsubsection{##1}{##2}\fi}
    \renewcommand{\l@paragraph}[2]{\ifcvpr@inappendix \cvpr@orig@paragraph{##1}{##2}\fi}
    \renewcommand{\l@subparagraph}[2]{\ifcvpr@inappendix \cvpr@orig@subparagraph{##1}{##2}\fi}
    \renewcommand{\CVPRAppendixTOCMarker}{\global\cvpr@inappendixtrue}
    \tableofcontents
  \endgroup
}
\makeatother

\definecolor{mri}{RGB}{66,133,244}    % Blue-ish for MRI
\definecolor{ct}{RGB}{220,53,69}      % Red-ish for CT
\definecolor{pet}{RGB}{255,140,0}     % Orange for PET

\definecolor{goodred}{HTML}{e67c73}
\definecolor{goodyellow}{HTML}{ffd666}
\definecolor{goodgreen}{HTML}{57bb8a}
\definecolor{usedblue}{HTML}{2a6fdb}
\definecolor{groupbg}{HTML}{eef1f5}
\usepackage{amssymb} % \blacktriangleleft

\pgfmathsetmacro{\minval}{0}  % Minimum value in the table
\pgfmathsetmacro{\maxval}{73}  % Maximum value in the table
\pgfmathsetmacro{\midval}{(\minval+\maxval)/2}  % Midpoint value for the yellow transition

\newcommand{\colorrange}[1]{%
    \pgfmathsetmacro{\PercentColor}{100.0*(#1-\minval)/(\maxval-\minval)}%
    \ifdim #1 pt < \minval pt
        \cellcolor{goodred}{#1}%
    \else
        \ifdim #1 pt > \maxval pt
            \cellcolor{goodgreen}{#1}%
        \else
            \ifdim #1 pt < \midval pt
                \pgfmathsetmacro{\PercentLow}{200.0*(#1-\minval)/(\maxval-\minval)}%
                \xdef\PercentLow{\PercentLow}%
                \cellcolor{goodyellow!\PercentLow!goodred}{#1}%
            \else
                \pgfmathsetmacro{\PercentHigh}{200.0*(#1-\midval)/(\maxval-\minval)}%
                \xdef\PercentHigh{\PercentHigh}%
                \cellcolor{goodgreen!\PercentHigh!goodyellow}{#1}%
            \fi
        \fi
    \fi
}

\pgfmathsetmacro{\minvala}{0}  % Minimum value in the table
\pgfmathsetmacro{\maxvala}{91}  % Maximum value in the table
\pgfmathsetmacro{\midvala}{(\minvala+\maxvala)/2}  % Midpoint value for the yellow transition

\newcommand{\colorrangea}[1]{%
    \pgfmathsetmacro{\PercentColor}{100.0*(#1-\minvala)/(\maxvala-\minvala)}%
    \ifdim #1 pt < \minvala pt
        \cellcolor{goodred}{#1}%
    \else
        \ifdim #1 pt > \maxvala pt
            \cellcolor{goodgreen}{#1}%
        \else
            \ifdim #1 pt < \midvala pt
                \pgfmathsetmacro{\PercentLow}{200.0*(#1-\minvala)/(\maxvala-\minvala)}%
                \xdef\PercentLow{\PercentLow}%
                \cellcolor{goodyellow!\PercentLow!goodred}{#1}%
            \else
                \pgfmathsetmacro{\PercentHigh}{200.0*(#1-\midvala)/(\maxvala-\minvala)}%
                \xdef\PercentHigh{\PercentHigh}%
                \cellcolor{goodgreen!\PercentHigh!goodyellow}{#1}%
            \fi
        \fi
    \fi
}

\pgfmathsetmacro{\minvalb}{0}  % Minimum value in the table
\pgfmathsetmacro{\maxvalb}{65}  % Maximum value in the table
\pgfmathsetmacro{\midvalb}{(\minvalb+\maxvalb)/2}  % Midpoint value for the yellow transition

\newcommand{\colorrangeb}[1]{%
    \pgfmathsetmacro{\PercentColor}{100.0*(#1-\minvalb)/(\maxvalb-\minvalb)}%
    \ifdim #1 pt < \minvalb pt
        \cellcolor{goodred}{#1}%
    \else
        \ifdim #1 pt > \maxvalb pt
            \cellcolor{goodgreen}{#1}%
        \else
            \ifdim #1 pt < \midvalb pt
                \pgfmathsetmacro{\PercentLow}{200.0*(#1-\minvalb)/(\maxvalb-\minvalb)}%
                \xdef\PercentLow{\PercentLow}%
                \cellcolor{goodyellow!\PercentLow!goodred}{#1}%
            \else
                \pgfmathsetmacro{\PercentHigh}{200.0*(#1-\midvalb)/(\maxvalb-\minvalb)}%
                \xdef\PercentHigh{\PercentHigh}%
                \cellcolor{goodgreen!\PercentHigh!goodyellow}{#1}%
            \fi
        \fi
    \fi
}

\pgfmathsetmacro{\minvalc}{0}  % Minimum value in the table
\pgfmathsetmacro{\maxvalc}{55}  % Maximum value in the table
\pgfmathsetmacro{\midvalc}{(\minvalc+\maxvalc)/2}  % Midpoint value for the yellow transition

\newcommand{\colorrangec}[1]{%
    \pgfmathsetmacro{\PercentColor}{100.0*(#1-\minvalc)/(\maxvalc-\minvalc)}%
    \ifdim #1 pt < \minvalc pt
        \cellcolor{goodred}{#1}%
    \else
        \ifdim #1 pt > \maxvalc pt
            \cellcolor{goodgreen}{#1}%
        \else
            \ifdim #1 pt < \midvalc pt
                \pgfmathsetmacro{\PercentLow}{200.0*(#1-\minvalc)/(\maxvalc-\minvalc)}%
                \xdef\PercentLow{\PercentLow}%
                \cellcolor{goodyellow!\PercentLow!goodred}{#1}%
            \else
                \pgfmathsetmacro{\PercentHigh}{200.0*(#1-\midvalc)/(\maxvalc-\minvalc)}%
                \xdef\PercentHigh{\PercentHigh}%
                \cellcolor{goodgreen!\PercentHigh!goodyellow}{#1}%
            \fi
        \fi
    \fi
}

\pgfmathsetmacro{\minvald}{0}  % Minimum value in the table
\pgfmathsetmacro{\maxvald}{53}  % Maximum value in the table
\pgfmathsetmacro{\midvald}{(\minvald+\maxvald)/2}  % Midpoint value for the yellow transition

\newcommand{\colorranged}[1]{%
    \pgfmathsetmacro{\PercentColor}{100.0*(#1-\minvald)/(\maxvald-\minvald)}%
    \ifdim #1 pt < \minvald pt
        \cellcolor{goodred}{#1}%
    \else
        \ifdim #1 pt > \maxvald pt
            \cellcolor{goodgreen}{#1}%
        \else
            \ifdim #1 pt < \midvald pt
                \pgfmathsetmacro{\PercentLow}{200.0*(#1-\minvald)/(\maxvald-\minvald)}%
                \xdef\PercentLow{\PercentLow}%
                \cellcolor{goodyellow!\PercentLow!goodred}{#1}%
            \else
                \pgfmathsetmacro{\PercentHigh}{200.0*(#1-\midvald)/(\maxvald-\minvald)}%
                \xdef\PercentHigh{\PercentHigh}%
                \cellcolor{goodgreen!\PercentHigh!goodyellow}{#1}%
            \fi
        \fi
    \fi
}

\pgfmathsetmacro{\minvale}{0}  % Minimum value in the table
\pgfmathsetmacro{\maxvale}{84}  % Maximum value in the table
\pgfmathsetmacro{\midvale}{(\minvale+\maxvale)/2}  % Midpoint value for the yellow transition

\newcommand{\colorrangee}[1]{%
    \pgfmathsetmacro{\PercentColor}{100.0*(#1-\minvale)/(\maxvale-\minvale)}%
    \ifdim #1 pt < \minvale pt
        \cellcolor{goodred}{#1}%
    \else
        \ifdim #1 pt > \maxvale pt
            \cellcolor{goodgreen}{#1}%
        \else
            \ifdim #1 pt < \midvale pt
                \pgfmathsetmacro{\PercentLow}{200.0*(#1-\minvale)/(\maxvale-\minvale)}%
                \xdef\PercentLow{\PercentLow}%
                \cellcolor{goodyellow!\PercentLow!goodred}{#1}%
            \else
                \pgfmathsetmacro{\PercentHigh}{200.0*(#1-\midvale)/(\maxvale-\minvale)}%
                \xdef\PercentHigh{\PercentHigh}%
                \cellcolor{goodgreen!\PercentHigh!goodyellow}{#1}%
            \fi
        \fi
    \fi
}

\pgfmathsetmacro{\minvalz}{0}  % Minimum value in the table
\pgfmathsetmacro{\maxvalz}{49}  % Maximum value in the table
\pgfmathsetmacro{\midvalz}{(\minvalz+\maxvalz)/2}  % Midpoint value for the yellow transition

\newcommand{\colorrangez}[1]{%
    \pgfmathsetmacro{\PercentColor}{100.0*(#1-\minvalz)/(\maxvalz-\minvalz)}%
    \ifdim #1 pt < \minvalz pt
        \cellcolor{goodred}{#1}%
    \else
        \ifdim #1 pt > \maxvalz pt
            \cellcolor{goodgreen}{#1}%
        \else
            \ifdim #1 pt < \midvalz pt
                \pgfmathsetmacro{\PercentLow}{200.0*(#1-\minvalz)/(\maxvalz-\minvalz)}%
                \xdef\PercentLow{\PercentLow}%
                \cellcolor{goodyellow!\PercentLow!goodred}{#1}%
            \else
                \pgfmathsetmacro{\PercentHigh}{200.0*(#1-\midvalz)/(\maxvalz-\minvalz)}%
                \xdef\PercentHigh{\PercentHigh}%
                \cellcolor{goodgreen!\PercentHigh!goodyellow}{#1}%
            \fi
        \fi
    \fi
}

\pgfmathsetmacro{\minvalq}{0}  % Minimum value in the table
\pgfmathsetmacro{\maxvalq}{40}  % Maximum value in the table
\pgfmathsetmacro{\midvalq}{(\minvalq+\maxvalq)/2}  % Midpoint value for the yellow transition

\newcommand{\colorrangeq}[1]{%
    \pgfmathsetmacro{\PercentColor}{100.0*(#1-\minvalq)/(\maxvalq-\minvalq)}%
    \ifdim #1 pt < \minvalq pt
        \cellcolor{goodred}{#1}%
    \else
        \ifdim #1 pt > \maxvalq pt
            \cellcolor{goodgreen}{#1}%
        \else
            \ifdim #1 pt < \midvalq pt
                \pgfmathsetmacro{\PercentLow}{200.0*(#1-\minvalq)/(\maxvalq-\minvalq)}%
                \xdef\PercentLow{\PercentLow}%
                \cellcolor{goodyellow!\PercentLow!goodred}{#1}%
            \else
                \pgfmathsetmacro{\PercentHigh}{200.0*(#1-\midvalq)/(\maxvalq-\minvalq)}%
                \xdef\PercentHigh{\PercentHigh}%
                \cellcolor{goodgreen!\PercentHigh!goodyellow}{#1}%
            \fi
        \fi
    \fi
}

\pgfmathsetmacro{\minvalr}{0}  % Minimum value in the table
\pgfmathsetmacro{\maxvalr}{42}  % Maximum value in the table
\pgfmathsetmacro{\midvalr}{(\minvalr+\maxvalr)/2}  % Midpoint value for the yellow transition

\newcommand{\colorranger}[1]{%
    \pgfmathsetmacro{\PercentColor}{100.0*(#1-\minvalr)/(\maxvalr-\minvalr)}%
    \ifdim #1 pt < \minvalr pt
        \cellcolor{goodred}{#1}%
    \else
        \ifdim #1 pt > \maxvalr pt
            \cellcolor{goodgreen}{#1}%
        \else
            \ifdim #1 pt < \midvalr pt
                \pgfmathsetmacro{\PercentLow}{200.0*(#1-\minvalr)/(\maxvalr-\minvalr)}%
                \xdef\PercentLow{\PercentLow}%
                \cellcolor{goodyellow!\PercentLow!goodred}{#1}%
            \else
                \pgfmathsetmacro{\PercentHigh}{200.0*(#1-\midvalr)/(\maxvalr-\minvalr)}%
                \xdef\PercentHigh{\PercentHigh}%
                \cellcolor{goodgreen!\PercentHigh!goodyellow}{#1}%
            \fi
        \fi
    \fi
}

\pgfmathsetmacro{\minvals}{0}  % Minimum value in the table
\pgfmathsetmacro{\maxvals}{30}  % Maximum value in the table
\pgfmathsetmacro{\midvals}{(\minvals+\maxvals)/2}  % Midpoint value for the yellow transition

\newcommand{\colorranges}[1]{%
    \pgfmathsetmacro{\PercentColor}{100.0*(#1-\minvals)/(\maxvals-\minvals)}%
    \ifdim #1 pt < \minvals pt
        \cellcolor{goodred}{#1}%
    \else
        \ifdim #1 pt > \maxvals pt
            \cellcolor{goodgreen}{#1}%
        \else
            \ifdim #1 pt < \midvals pt
                \pgfmathsetmacro{\PercentLow}{200.0*(#1-\minvals)/(\maxvals-\minvals)}%
                \xdef\PercentLow{\PercentLow}%
                \cellcolor{goodyellow!\PercentLow!goodred}{#1}%
            \else
                \pgfmathsetmacro{\PercentHigh}{200.0*(#1-\midvals)/(\maxvals-\minvals)}%
                \xdef\PercentHigh{\PercentHigh}%
                \cellcolor{goodgreen!\PercentHigh!goodyellow}{#1}%
            \fi
        \fi
    \fi
}

\usepackage{graphicx,verbatim}
\usepackage{booktabs}
\usepackage{amsfonts}
\usepackage{amsmath}
\usepackage{bm}
\usepackage{booktabs}
\usepackage{makecell}
\usepackage{caption}
\usepackage{subcaption}
\usepackage[hidelinks]{hyperref}
\usepackage{pifont}
\usepackage{adjustbox}

\begin{document}
\title{Anguinus Sculpturae: Compositional Synthesis of Peak-Enhancement Breast DCE-MRI Scans}
\titlerunning{Anguinus Sculpturae}

\author{Benjamin Hamm\inst{1,4} %index{Hamm, Benjamin}
\and Nico Albert Disch\inst{1,2,3} %index{Disch, Nico Albert}
\and Maximilian Rokuss\inst{1,2,3,5} %index{Rokuss, Maximilian}
\and Yannick Kirchhoff\inst{1,2,3} %index{Kirchhoff, Yannick}
\and Constantin Ulrich\inst{1} %index{Ulrich, Constantin}
\and Klaus Maier-Hein\inst{1,4,6} %index{Maier-Hein, Klaus}
}
\authorrunning{Hamm et al.}
\institute{German Cancer Research Center (DKFZ) Heidelberg, Division of Medical Image Computing, Germany
\and
Faculty of Mathematics and Computer Science, Heidelberg University, Germany
\and
HIDSS4Health -- Helmholtz Information and Data Science School for Health, Karlsruhe/Heidelberg, Germany
\and
Medical Faculty, Heidelberg University, Germany
\and
Helmholtz Imaging, German Cancer Research Center (DKFZ), Heidelberg, Germany
\and
Pattern Analysis and Learning Group, Department of Radiation Oncology, Heidelberg University Hospital, Germany\\
    \email{benjamin.hamm@dkfz-heidelberg.de}}
  
\maketitle

\begin{abstract}

Dynamic contrast-enhanced breast MRI (DCE-MRI) is rich in anatomical and perfusion
information, but its reliance on gadolinium-based contrast agents raises safety concerns and
adds cost. Virtual contrast enhancement, synthesizing post-contrast from pre-contrast images,
is a promising alternative. We address the MAMA-SYNTH challenge task of predicting
peak-enhancement breast MRI. Rather than adopting the full machinery of diffusion or flow
matching, we observe that under a rectified, straight-line path the generative process
collapses to a single difference prediction: the synthetic peak image is the pre-contrast image
plus a predicted enhancement map, recovered in one forward pass. Around this we build
\emph{Anguinus Sculpturae}, a compositional pipeline in which nnU-Net segmentations of lesion,
foreground and breast region guide two generators---one optimized for global fidelity, one for
lesion structure through an asymmetric Tversky term routed via a frozen segmenter---composited
region-wise with Gaussian-weighted blending. On the held-out Duke subset of MAMA-MIA our model
achieves the best FRD and Dice among all evaluated
variants, showing that single-step difference prediction with segmentation guidance suffices to
recover both global fidelity and lesion structure. Code is available at
\url{https://github.com/MIC-DKFZ/AnguinusSculpturae}.


\end{abstract}

\section{Introduction}
Breast cancer remains among the foremost contributors to cancer death in
women~\cite{sung2021global}. Dynamic contrast-enhanced magnetic resonance imaging (DCE-MRI) is
particularly valuable for its assessment, combining detailed anatomy with temporal information
on tissue perfusion and vascular permeability~\cite{mann2019breast}; in high-risk screening it
reaches markedly higher sensitivity than mammography and ultrasound, regardless of breast
density~\cite{riedl2015triple}, and it increasingly feeds automated analysis such as breast cancer
classification~\cite{hamm2025meisenmeister}. Because it hinges on injecting gadolinium-based contrast agents,
however, it raises patient-safety concerns, excludes certain individuals, and adds cost and procedural 
burden~\cite{marckmann2006nephrogenic,olchowy2017presence,idee2006clinical}. Prompted by this, a 
growing body of work has established that post-contrast breast MRI can be inferred from pre-contrast
acquisitions~\cite{mullerfranzes2023using,osuala2024pre,ibarra2025comparing}, positioning virtual
contrast enhancement as a substitute for, or complement to, conventional DCE-MRI. The MAMA-SYNTH
challenge~\cite{osuala2026mamasynth} offers a standardized, clinically grounded benchmark for
the task; this paper reports the method we submitted to it.

Work on this task has relied heavily on generative adversarial networks
(GANs)~\cite{goodfellow2014generative,isola2017image}. In breast MRI, M\"uller-Franzes et
al.~\cite{mullerfranzes2023using} recover contrast-enhanced images from unenhanced T1- and
T2-weighted acquisitions and from simulated low-dose ones, and Osuala et
al.~\cite{osuala2024pre,osuala2025simulating} synthesize post- from pre-contrast images, showing
the result supports downstream tumor segmentation. Such models inherit the training signal they
are built on: adversarial objectives are hard to stabilize, vulnerable to mode collapse, and can
encourage hallucinated anatomy, which is problematic where reliability is essential. Liebert et
al.~\cite{liebert2025impact}, comparing which non-contrast input sequences the task needs, chose
a plain encoder--decoder over a GAN for its robustness.

Diffusion models trade that min-max game for a regression objective, generating an image by
gradually denoising a Gaussian sample through a learned reverse process~\cite{ho2020denoising},
and the substitution has since been carried to this task: a latent diffusion model conditioned on
acquisition time can approximate contrast kinetics~\cite{osuala2024towards}, and Ibarra et
al.~\cite{ibarra2025comparing} compare conditional diffusion models on the same pre-to-post
problem. The stability is paid for at inference, by sampling that can run to a thousand
sequential steps~\cite{ho2020denoising}. Flow matching~\cite{lipman2022flow,liu2022flow} trims
that cost, regressing a velocity field $v_\theta(x_t,t)$ that transports a source onto a target
distribution along a prescribed path, samples following from integrating
$\dot{x}_t=v_\theta(x_t,t)$ over $t\in[0,1]$. The apparatus, though, is the same in kind---a time
variable, sampling along the path during training, multi-step integration at inference---and for
this task none of it is needed.

What lets us drop it is a second observation from that same work. Ibarra et
al.~\cite{ibarra2025comparing} report that predicting the \emph{subtraction} image consistently
beats predicting the post-contrast image directly, and under the rectified (straight-line) path
$x_t=(1-t)\,x_0+t\,x_1$~\cite{liu2022flow} that target is not merely a better parameterization
but the velocity field itself: the path has constant velocity
$v=x_1-x_0$, so with $x_0=\mathrm{pre}$ and $x_1=\mathrm{peak}$ the field a flow-matching model
would fit is exactly the subtraction (contrast-enhancement) image, independent of $t$. Each part
of the apparatus then falls away in turn. A field that does not depend on $t$ needs no time
conditioning and no sampling along the path, and traversing a straight path at constant speed
needs a single Euler step with $\Delta t=1$, which recovers the target exactly,
$\mathrm{pre}+\mathrm{sub}=\mathrm{peak}$---the reasoning that one-step rectified-flow models such
as InstaFlow~\cite{liu2023instaflow} reach by straightening their trajectories, and that this task
supplies for free. We therefore predict $\mathrm{sub}$ from the pre-contrast image in one forward
pass and take $\mathrm{pre}+\mathrm{sub}$ as the synthetic peak image: a plain regression network,
arriving by way of flow matching where Liebert et al.~\cite{liebert2025impact} were already
heading.

\section{Methods}

\subsection{Data}\label{sub:data}

We first expanded the training pool beyond MAMA-MIA~\cite{garrucho2025large} with public
data. Three further cohorts provide lesion segmentations: Yunnan~\cite{zhang2023breastcancer},
TCGA-Breast-Radiogenomics~\cite{morris2014breastmri} and the QIN Breast DCE-MRI
collection~\cite{huang2014qinbreastdce}. We then added pseudo-labels: using the
nnU-Net weights released with MAMA-MIA we predicted lesion masks for the patients not covered by
MAMA-MIA within Duke~\cite{duke}, ISPY1~\cite{ispy1}, ISPY2~\cite{ispy2} and
NACT~\cite{newitt2016singlebreastdce}, and for follow-up visits of patients already included.
Being from the same source cohorts, these are strongly in-domain and we expect their
pseudo-labels to be reliable. Beyond them we added seven public cohorts carrying only
breast-level malignancy status (ACRIN 6667 contralateral, AMBL, ODELIA~\cite{odelia},
BREAST-DIAG\-NOSIS, fastMRI Breast~\cite{solomon2025fastmribreast}, QIN-BREAST and
QIN-BREAST-02), pseudo-labeled the same
way; source, DOI and licence for each are documented in our
repository.\footnote{\url{https://github.com/MIC-DKFZ/AnguinusSculpturae}}
Foreground and breast-region pseudo-labels come from publicly available
models~\cite{nohel2025unified,rokuss2025breastdivider}. Every
pseudo-label is one automatic forward pass with no manual correction at any stage, and we
validated none of them, assuming they would be accurate enough.

The pool holds 7{,}555 cases (799{,}563 slices), split into four
\emph{leave-one-cohort-out} folds defined by the four MAMA-MIA acquisition centers so that
validation always measures cross-site generalization: fold 0 holds out the 291 MAMA-MIA Duke
cases, folds 1--3 hold out NACT (with QIN Breast DCE-MRI), ISPY1 and ISPY2. Every cohort above
enters the training half of every fold but the one held out, so the additional data are extra
training material rather than extra ensemble members---the ensemble has exactly one member per
fold. Fold 0, held out identically in pre-training, is the development set for all numbers in
Sec.~\ref{sec:results}.

\subsection{Pretraining}

We performed a standard masked autoencoder (MAE) pretraining, closely
following~\cite{wald2025revisiting} but in 2D, on 53{,}168 volumes (1000 epochs, initial
learning rate $1\times10^{-4}$, reconstruction loss only). Here we did not restrict the input
to the pre-contrast and peak phases but used all available phases. The resulting weights
initialize both generators; the three segmenters are trained from scratch.

\subsection{Anguinus Sculpturae}\label{sub:as}

\begin{figure}[!t]
 \centering
 \includegraphics[width=0.76\linewidth]{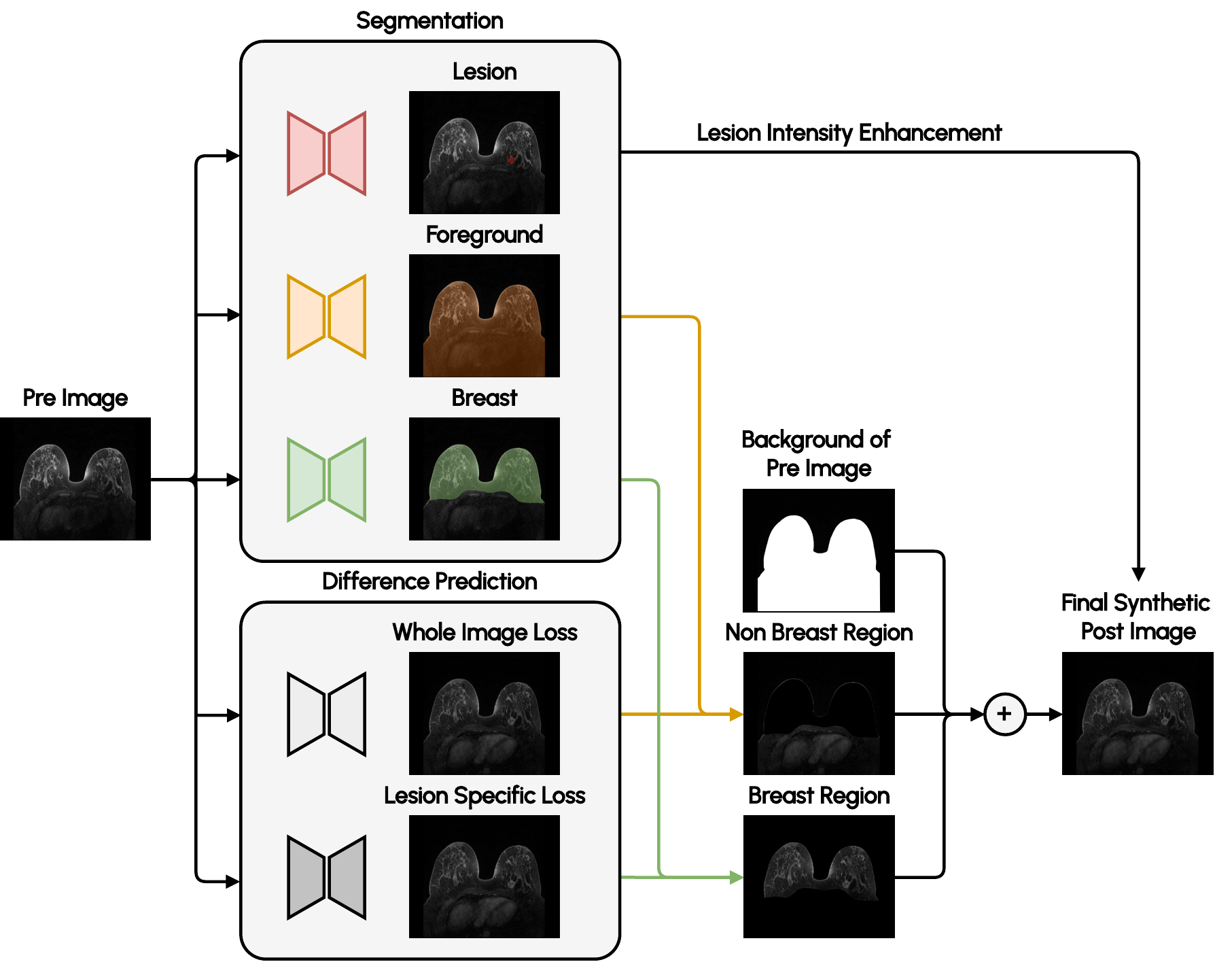}
 \caption{The \emph{Anguinus Sculpturae} pipeline. From a pre-contrast image, three 2D
 nnU-Nets segment lesion, foreground and breast region (top) while two generators predict the
 contrast enhancement, one with a whole-image loss and one with lesion-specific losses
 (bottom). The output is composited region-wise with Gaussian-weighted blending: background
 from the pre-contrast image, non-breast foreground from the whole-image-loss model, breast
 region from the lesion-loss model.}
 \label{fig:as}
\end{figure}

\noindent\textbf{Networks.} All five networks share one backbone, an unmodified 2D
\texttt{Residual\allowbreak Encoder\allowbreak UNet} from
nnU-Net~\cite{isensee2021nnu}: $512\times512$ input, seven stages with 32, 64,
128, 256, 512, 512 and 512 features, $3\times3$ kernels, 1, 3, 4, 6, 6, 6 and 6 residual blocks
per encoder stage, one convolution per decoder stage, instance normalization, leaky ReLU and
$z$-score normalization. The segmenters keep the softmax head and
nnU-Net's deep-supervised Dice-plus-cross-entropy loss; the generators replace it with a
single-channel linear head regressing $\mathrm{sub}$ at full resolution, without deep
supervision.

\noindent\textbf{Guidance.} Given a pre-contrast image we compute three 2D
segmentations---foreground mask, breast region, lesion---as guidance signals
(Fig.~\ref{fig:as}).
The lesion segmenter learns from masks annotated on contrast-enhanced images, carrying annotations
across acquisition techniques as done for brain metastases~\cite{wald2025enhancing}. Because the lesion
segmenter can return an empty mask, we lower its decision threshold per slice \emph{at
inference}, in steps of $0.05$ from the default $0.5$, until at least one voxel inside the
breast region is labeled lesion, ensuring a non-trivial lesion signal in every case: a test-time
search over the threshold of a fixed network, playing no part in training.

\noindent\textbf{Objectives.} We synthesize the peak-enhancement image under two loss
configurations,
\begin{equation}\label{eq:losses}
\mathcal{L}_{\mathrm{img}}=\mathcal{L}_{\mathrm{MSE}}+\mathcal{L}_{\mathrm{LPIPS}},
\qquad
\mathcal{L}_{\mathrm{les}}=\mathcal{L}_{\mathrm{MSE}}+\tfrac{1}{2}\mathcal{L}_{\mathrm{LPIPS}}
+\mathcal{L}^{\mathrm{ROI}}_{\mathrm{SSIM}}
+\mathcal{L}^{\mathrm{seg}}_{\mathrm{Dice}}+\mathcal{L}^{\mathrm{seg}}_{\mathrm{Tversky}},
\end{equation}
where $\mathcal{L}_{\mathrm{MSE}}$ is taken over acquired pixels only and
$\mathcal{L}_{\mathrm{LPIPS}}$~\cite{zhang2018unreasonable} uses the standard
ImageNet-pretrained AlexNet backbone on slices clipped at $\pm5\sigma$;
$\mathcal{L}^{\mathrm{ROI}}_{\mathrm{SSIM}}=1-\mathrm{SSIM}$ comes from the full-image SSIM map
($7\times7$ window, data range $10$ in $z$-score units) averaged within the lesion ROI. The last
two terms form a segmentation-consistency loss: the frozen lesion segmenter provided by the
organizers is applied to the \emph{synthesized} slice and its lesion probability scored against
the reference mask, gradients passing back through the frozen segmenter into the generator.
That segmenter is used only in training, never at inference. We deliberately make the Tversky
term $\mathrm{TP}/(\mathrm{TP}+\alpha\,\mathrm{FP}+\beta\,\mathrm{FN})$ asymmetric, at
$\alpha=0.2$, $\beta=0.8$: a missed lesion voxel counts four times a false-positive one,
encouraging unambiguously detectable lesions. All weights are $1$ except LPIPS in the lesion
generator, halved because at $1$ the perceptual term dominated the lesion terms early in
training. The $4{:}1$ asymmetry was fixed a priori and, like the other weights, never
swept---within the challenge timeline we ran no systematic hyperparameter search, and the
ablations in Sec.~\ref{sec:results} vary one component at a time rather than tune it.

\noindent\textbf{Enhancement magnitude.} The lesion-region objective was meant to combine two
complementary terms. $\mathcal{L}^{\mathrm{ROI}}_{\mathrm{SSIM}}$ is computed on $z$-scored
intensities and so is invariant to any affine transformation $aX+b$: it constrains the structure
and texture of the enhancement but leaves its absolute brightness and contrast free. MSE does not pin
those down either. It is averaged over the whole slice, where the lesion is well under two percent
of the pixels, so its amplitude there is nearly free to be wrong; and the L2 optimum is the
conditional mean $\mathbb{E}[\mathrm{sub}\mid\mathrm{pre}]$, a hedge over how strongly \emph{this}
lesion will enhance---precisely what the pre-contrast image does not reveal. That distribution is
right-skewed and conspicuity lives in its upper tail, so the hedge renders strongly enhancing
lesions too weakly, and the matching uncertainty about lesion extent spreads the predicted
enhancement over a larger area, lowering its peak further. MSE
compresses the dynamic range of the enhancement rather than biasing it one way everywhere, but the
cases it renders too weakly are the ones that matter. We therefore sought a
second lesion-region term recovering what the invariant losses discard---the absolute offset and
scale. We could not make that objective train stably,
and under the challenge timeframe adopted a simpler alternative: we scale intensities inside the
predicted lesion region by $1.25$, a 25\% relative boost, applied multiplicatively so the
lesion's texture is preserved, through a mask feathered with a Gaussian of $\sigma=2$\,px. The
value was not swept: we tried $1.25$, it worked, and the challenge timeline left no room to try
others.

\noindent\textbf{Composition.} Finally we assemble the output from three sources: the air
region (the non-foreground prediction) from the pre-contrast image, the breast region from the
lesion-loss-guided model, the remaining foreground from the global-fidelity model, with
Gaussian-weighted blending ($\sigma=4$\,px) at the boundaries. Copying the air region is
deliberate: outside the tissue support there is no enhancement to predict, so a generator can
only add noise there. On the development fold this improved FRD and left the other metrics
essentially unchanged, as one expects of a region contributing error but no signal.

\noindent\textbf{Training and inference.} The generators start from the MAE checkpoint and are
fine-tuned at batch size 48 with a linear warm-up over 50 epochs from $2\times10^{-5}$ to a peak
learning rate of $1\times10^{-3}$, then PolyLR decay over 150 epochs (lesion loss) and 200
(whole-image loss), weight decay $1\times10^{-4}$. The segmenters are trained from scratch on
nnU-Net's default schedule (PolyLR from $1\times10^{-2}$, no warm-up): lesion at batch size 64
for 280 epochs on fold 0 and 1000 on folds 1--3, breast and foreground for 100 and 77 epochs at
batch size 32, two runs stopped by hand once validation had plateaued. Checkpoints were selected by best
validation performance---lowest loss for the global-fidelity model, best Dice for the
lesion-objective model---and at inference we apply test-time augmentation over all axis-aligned
flips. The submitted model ensembles the four folds of the lesion segmenter and of both
generators, one per MAMA-MIA acquisition center; breast and foreground segmentation proved easy
enough that per-fold models added nothing, so one fold-0 model serves for each: 14 trained
networks, five of which run per slice.

\noindent\textbf{Ablated components.} Four components appear in Sec.~\ref{sec:results} but not in
the final pipeline. \emph{GAN} adds a patch discriminator on the synthetic slice with an
adversarial term at weight $0.02$; \emph{deep supervision} adds auxiliary reconstruction losses on
the $256^2$ and $128^2$ decoder outputs (weights $0.25$, $0.125$); \emph{resize conv} replaces the
decoder's transposed convolutions with nearest-neighbor upsampling plus a convolution;
\emph{Curia feature loss} adds a feature-space term computed with the Curia radiology
foundation model~\cite{dancette2025curia}.

\section{Experiments and Results}\label{sec:results}

\begin{table}[t]
\centering
\caption{Ablations on the held-out Duke part of MAMA-MIA (fold 0, 291 cases); every row in
blocks A--C is a single fold-0 network, or in C the composition of two, with flip test-time
augmentation. Arrows give the direction of improvement; \textbf{best} and \underline{second-best}
per metric over rows 1--12, cells shaded per column from worst (red) to best (green). An unindented
``+'' row changes one component of the first row of its block; an indented row changes one
component of the nearest unindented row above it.
\textcolor{usedblue}{$\blacktriangleleft$} marks the configurations used in the final pipeline.
The whole-image-loss generator gets no lesion supervision and never supplies the breast region, so
it is scored on fidelity only~(--). Block D is the challenge validation leaderboard: external test
data, the submitted four-fold ensemble, shaded on the same scale but not ranked and not comparable
to A--C.}
\renewcommand{\arraystretch}{1.1}
\setlength{\tabcolsep}{4pt}
\adjustbox{width=\textwidth}{
% Generated by presentation/make_ablation_table.py -- regenerate there, then paste.
\begin{tabular}{r l|cc|cc|cc|cc}
\toprule
\# & Configuration & MSE\,$\downarrow$ & LPIPS\,$\downarrow$ & SSIM$_\mathrm{tum}$\,$\uparrow$ & FRD\,$\downarrow$ & Dice\,$\uparrow$ & HD$_{95}$\,$\downarrow$ & AUROC$_\mathrm{con}$\,$\uparrow$ & AUROC$_\mathrm{ROI}$\,$\uparrow$ \\
\midrule
\rowcolor{groupbg}\multicolumn{10}{l}{\textbf{A\quad Lesion-loss generator}\enspace{\small single network \textperiodcentered\ base loss: MSE + $\tfrac12$LPIPS + SSIM$_\mathrm{ROI}$ + Dice$_\mathrm{seg}$}}\\
\textcolor{black!50}{\small 1} & Base & \cellcolor{goodgreen!46!goodyellow}0.784 & \cellcolor{goodyellow!47!goodred}0.140 & \cellcolor{goodgreen!64!goodyellow}0.474 & \cellcolor{goodgreen!54!goodyellow}\underline{25.04} & \cellcolor{goodyellow!64!goodred}0.578 & \cellcolor{goodyellow!92!goodred}98.0 & \cellcolor{goodgreen!14!goodyellow}0.936 & \cellcolor{goodgreen!15!goodyellow}0.626 \\
\textcolor{black!50}{\small 2} & + Curia feature loss & \cellcolor{goodyellow!90!goodred}0.812 & \cellcolor{goodyellow!73!goodred}0.136 & \cellcolor{goodgreen!66!goodyellow}0.475 & \cellcolor{goodyellow!20!goodred}27.72 & \cellcolor{goodyellow!44!goodred}0.568 & \cellcolor{goodyellow!76!goodred}102.9 & \cellcolor{goodgreen!30!goodyellow}0.940 & \cellcolor{goodgreen!66!goodyellow}\underline{0.641} \\
\textcolor{black!50}{\small 3} & \hspace{1.1em}+ GAN discriminator & \cellcolor{goodyellow!90!goodred}0.812 & \cellcolor{goodgreen!7!goodyellow}0.131 & \cellcolor{goodgreen!70!goodyellow}0.477 & \cellcolor{goodyellow!25!goodred}27.61 & \cellcolor{goodyellow!38!goodred}0.565 & \cellcolor{goodyellow!28!goodred}117.3 & \cellcolor{goodgreen!14!goodyellow}0.936 & \cellcolor{goodgreen!100!goodyellow}\textbf{0.651} \\
\textcolor{black!50}{\small 4} & \hspace{1.1em}+ deep supervision & \cellcolor{goodyellow!96!goodred}0.809 & \cellcolor{goodgreen!7!goodyellow}0.131 & \cellcolor{goodgreen!58!goodyellow}0.471 & \cellcolor{goodyellow!41!goodred}27.28 & \cellcolor{goodyellow!12!goodred}0.552 & \cellcolor{goodyellow!23!goodred}118.7 & \cellcolor{goodgreen!38!goodyellow}\textbf{0.942} & \cellcolor{goodgreen!32!goodyellow}0.631 \\
\textcolor{black!50}{\small 5} & \hspace{1.1em}+ resize-conv decoder & \cellcolor{goodyellow!42!goodred}0.836 & \cellcolor{goodyellow!67!goodred}0.137 & \cellcolor{goodgreen!55!goodyellow}0.469 & \cellcolor{goodgreen!15!goodyellow}25.81 & \cellcolor{goodyellow!36!goodred}0.564 & \cellcolor{goodyellow!4!goodred}124.4 & \cellcolor{goodgreen!26!goodyellow}0.939 & \cellcolor{goodgreen!66!goodyellow}\underline{0.641} \\
\textcolor{black!50}{\small 6} & + Tversky term ($\alpha{=}0.2,\ \beta{=}0.8$) & \cellcolor{goodyellow!64!goodred}0.825 & \cellcolor{goodyellow!0!goodred}0.147 & \cellcolor{goodgreen!72!goodyellow}0.478 & \cellcolor{goodyellow!92!goodred}26.26 & \cellcolor{goodgreen!16!goodyellow}0.604 & \cellcolor{goodgreen!10!goodyellow}92.5 & \cellcolor{goodyellow!62!goodred}0.923 & \cellcolor{goodyellow!37!goodred}0.603 \\
\textcolor{black!50}{\small 7} & \hspace{1.1em}+ MAE pre-training\,\textcolor{usedblue}{$\blacktriangleleft$} & \cellcolor{goodgreen!38!goodyellow}0.788 & \cellcolor{goodyellow!40!goodred}0.141 & \cellcolor{goodgreen!100!goodyellow}\textbf{0.493} & \cellcolor{goodyellow!29!goodred}27.54 & \cellcolor{goodgreen!86!goodyellow}\underline{0.639} & \cellcolor{goodgreen!96!goodyellow}\textbf{66.8} & \cellcolor{goodyellow!98!goodred}0.932 & \cellcolor{goodyellow!92!goodred}0.619 \\
\midrule
\rowcolor{groupbg}\multicolumn{10}{l}{\textbf{B\quad Whole-image-loss generator}\enspace{\small single network \textperiodcentered\ loss: MSE + LPIPS \textperiodcentered\ no lesion supervision}}\\
\textcolor{black!50}{\small 8} & From scratch (batch 32) & \cellcolor{goodgreen!10!goodyellow}0.802 & \cellcolor{goodgreen!100!goodyellow}\textbf{0.117} & \cellcolor{goodyellow!0!goodred}0.387 & \cellcolor{goodyellow!89!goodred}26.32 & \textcolor{black!40}{--} & \textcolor{black!40}{--} & \textcolor{black!40}{--} & \textcolor{black!40}{--} \\
\textcolor{black!50}{\small 9} & + MAE pre-training & \cellcolor{goodgreen!46!goodyellow}0.784 & \cellcolor{goodgreen!87!goodyellow}\underline{0.119} & \cellcolor{goodyellow!51!goodred}0.414 & \cellcolor{goodgreen!23!goodyellow}25.65 & \textcolor{black!40}{--} & \textcolor{black!40}{--} & \textcolor{black!40}{--} & \textcolor{black!40}{--} \\
\textcolor{black!50}{\small 10} & \hspace{1.1em}+ batch 48, 200 epochs\,\textcolor{usedblue}{$\blacktriangleleft$} & \cellcolor{goodgreen!58!goodyellow}\textbf{0.778} & \cellcolor{goodgreen!67!goodyellow}0.122 & \cellcolor{goodyellow!62!goodred}0.420 & \cellcolor{goodyellow!33!goodred}27.44 & \textcolor{black!40}{--} & \textcolor{black!40}{--} & \textcolor{black!40}{--} & \textcolor{black!40}{--} \\
\midrule
\rowcolor{groupbg}\multicolumn{10}{l}{\textbf{C\quad Anguinus Sculpturae composition}\enspace{\small fold-0 generators of rows 7 and 10, masks from pre-contrast segmenters}}\\
\textcolor{black!50}{\small 11} & Breast $\leftarrow$ 7, other tissue $\leftarrow$ 10, air $\leftarrow$ pre & \cellcolor{goodgreen!56!goodyellow}\underline{0.779} & \cellcolor{goodgreen!20!goodyellow}0.129 & \cellcolor{goodgreen!100!goodyellow}\textbf{0.493} & \cellcolor{goodgreen!80!goodyellow}\textbf{24.50} & \cellcolor{goodgreen!68!goodyellow}0.630 & \cellcolor{goodgreen!81!goodyellow}71.3 & \cellcolor{goodyellow!98!goodred}0.932 & \cellcolor{goodyellow!0!goodred}0.592 \\
\textcolor{black!50}{\small 12} & + lesion boost $\times$1.25\,\textcolor{usedblue}{$\blacktriangleleft$} & \cellcolor{goodgreen!24!goodyellow}0.795 & \cellcolor{goodgreen!20!goodyellow}0.129 & \cellcolor{goodgreen!98!goodyellow}\underline{0.492} & \cellcolor{goodgreen!80!goodyellow}\textbf{24.50} & \cellcolor{goodgreen!88!goodyellow}\textbf{0.640} & \cellcolor{goodgreen!85!goodyellow}\underline{70.0} & \cellcolor{goodgreen!34!goodyellow}\underline{0.941} & \cellcolor{goodyellow!51!goodred}0.607 \\
\midrule[\heavyrulewidth]
\rowcolor{groupbg}\multicolumn{10}{l}{\textbf{D\quad Challenge validation leaderboard}\enspace{\small external test data \textperiodcentered\ submitted 4-fold ensemble of row 12 \textperiodcentered\ not comparable to A--C}}\\
 & \textbf{Anguinus Sculpturae} & \cellcolor{goodgreen!100!goodyellow}0.49 & \cellcolor{goodgreen!100!goodyellow}0.08 & \cellcolor{goodgreen!100!goodyellow}0.54 & \cellcolor{goodgreen!100!goodyellow}23.06 & \cellcolor{goodyellow!0!goodred}0.45 & \cellcolor{goodyellow!0!goodred}179.13 & \cellcolor{goodyellow!0!goodred}0.80 & \cellcolor{goodyellow!27!goodred}0.60 \\
\bottomrule
\end{tabular}
}
\label{tab:comparison}
\end{table}

\noindent\textbf{Protocol.} All numbers except the leaderboard row are means over the held-out Duke fold
(Sec.~\ref{sub:data}) under the challenge's eight metrics, computed with the organizers'
evaluation code. MSE and LPIPS~\cite{zhang2018unreasonable} score the whole image,
SSIM$_\mathrm{tum}$~\cite{wang2004image} is SSIM restricted to the tumor region, and
FRD~\cite{osuala2024towards,konz2025frd} a Fr\'echet distance over radiomic features. Dice and
HD$_{95}$ come from the organizers' frozen post-contrast lesion segmenter run on the synthetic
image against the reference mask. The AUROCs use pretrained radiomics classifiers:
AUROC$_\mathrm{con}$ separates synthetic post- from real pre-contrast images, while
AUROC$_\mathrm{ROI}$ separates the tumor ROI from a mirrored contralateral ROI of the same
synthetic image, measuring whether enhancement is specific to the lesion rather than spread
across the breast. Both AUROCs are cohort-level by construction---one value per model, no
per-case distribution. Development used this single held-out cohort throughout, without
significance testing, which limits how far any single-metric difference below should be read.

\begin{figure}[!t]
 \centering
 \includegraphics[width=\linewidth]{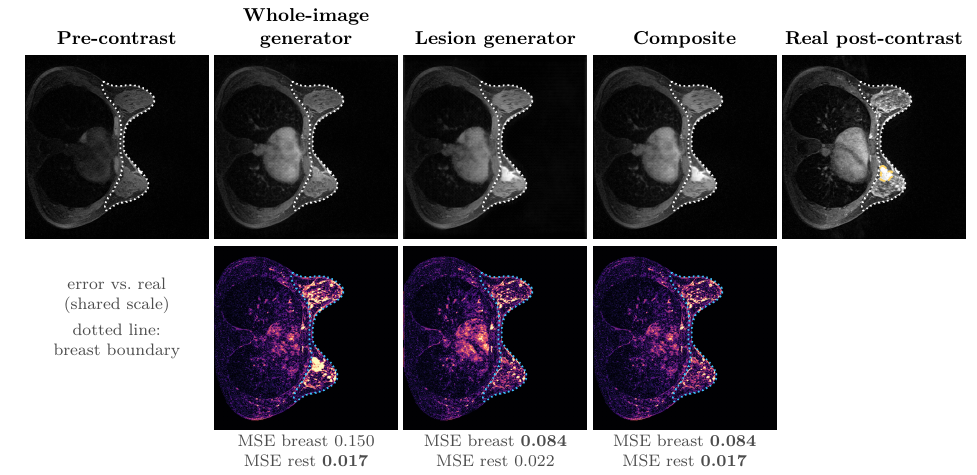}
 \caption{Why two generators, on Duke~799 (full slices), the case in which each generator most
 clearly wins its own region. Top: the generators' outputs, their composite (without boost) and
 the real image; bottom: error against the real image on one shared scale, the dotted line marking
 the breast boundary at which the outputs are composited. The composite keeps the lower error of
 each generator in its own region.}
 \label{fig:composition}
\end{figure}

\noindent\textbf{Findings.} The asymmetric Tversky term gave the strongest lesion overlap among
the single-component ablations, Dice $0.604$ at HD$_{95}$ $92.5$ against $0.552$--$0.568$ and
$103$--$124$ for the other variants. Above all, the two generators individually
traded one objective against the other---the whole-image-loss model strong on fidelity but weak on tumor structure
(SSIM$_\mathrm{tum}$ $0.420$), the lesion-specific model the reverse---while their region-wise
composition combined both (Fig.~\ref{fig:composition}), improving FRD
from about $27.5$ to $24.50$ over either alone. Deep supervision, resize convolutions and
the Curia feature loss~\cite{dancette2025curia} did not help. AUROC$_\mathrm{ROI}$ was hardest to improve: the
final model reaches $0.607$ while staying competitive on AUROC$_\mathrm{con}$ at $0.941$. The highest
value came with the GAN discriminator, at $0.651$, but it cost too much elsewhere
(MSE $0.812$).

\begin{figure}[!t]
 \centering
 \includegraphics[width=\linewidth]{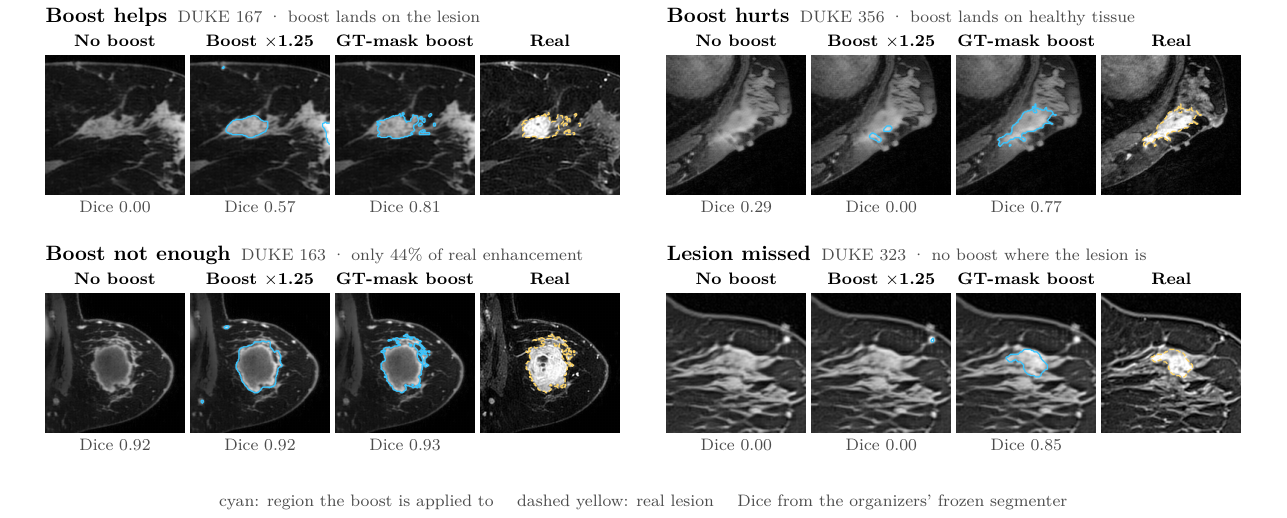}
 \caption{The lesion boost on four Duke cases: composite without boost, with the $\times1.25$
 boost inside the predicted lesion, with the boost inside the ground-truth lesion, and the real
 image. In Duke~167 the boost lands on the lesion, in Duke~356 on healthy tissue; in Duke~163 it
 is too weak, reaching 44\% of the real enhancement, and in Duke~323 the lesion prediction misses
 the lesion, so the boost is not applied where it is.}
 \label{fig:boost}
\end{figure}

\noindent\textbf{Mask dependence.} The boost is applied inside a \emph{predicted} mask, so the
frozen segmenter's errors reach the image as errors of extent rather than invented structure: the
boost is multiplicative and feathered, and rescales signal already present. An over-inclusive mask
therefore brightens tissue that should not enhance, as in Duke~356 (Fig.~\ref{fig:boost}), and a missed lesion leaves the boost unapplied,
as in Duke~323, where the enhancement the synthetic image shows comes from the generator alone. The
other frozen predictors enter the pipeline the same way, and a better-conditioned enhancement term would remove the dependence with the fixed boost.

\section{Discussion and Conclusion}
 
We approached the MAMA-SYNTH task by reducing it to its simplest sufficient form, a single
difference prediction $\mathrm{peak}=\mathrm{pre}+\mathrm{sub}$, and two design choices carried
that reduction. Region-wise composition let each generator specialize---the global-fidelity model
on the anatomy that dominates the image, the lesion-guided model on the small, low-cue enhancing
region that determines clinical value---and routing lesion supervision through a frozen segmenter
with an asymmetric Tversky term shifted the objective from looking realistic toward remaining
detectable.

The limitations point directly to future work. Enhancement magnitude is the clearest: the fixed
25\% boost is a crude stand-in for a principled term, and a better-conditioned formulation would
remove both the heuristic and the false-positive enhancement it can produce. The Tversky asymmetry
deserves the same treatment---it controls exactly the trade-off between missed and invented
lesions, and fixing it at $4{:}1$ without a sweep leaves that trade-off uncharacterized.
Finally, we leaned heavily on unvalidated pseudo-labels and used one
held-out cohort for both development and evaluation; multi-center validation with verified
annotations and reader assessment of hallucinated enhancement is the next step, for which
platforms that bring AI into clinical research environments~\cite{akunal2025kaapana} and
privacy-preserving federated training across sites~\cite{hamm2025efficient} offer a route.

One question this work sharpens: which perceptual objective to train on. LPIPS runs on an ImageNet
AlexNet that has never seen breast MRI, so a radiology foundation model like
Curia~\cite{dancette2025curia}, 2D and MR-exposed, should have been the better signal. It was
not---added as a feature loss it did not help. We would not read that as a
verdict on medical features. LPIPS is itself one of the challenge metrics, so no perceptual score
here is independent of it, and Woodland et al.~\cite{woodland2024feature} find ImageNet extractors
aligning with expert judgment better than domain-specific ones but test only supervised classifiers.
Whether medically pretrained features make a better perceptual loss is still untested.

\begin{credits}
\subsubsection{\ackname} This work was supported by the Helmholtz Association under the joint research program “HIDSS4Health – Helmholtz Information and Data Science School for Health” and under the Helmholtz Foundation Model Initiative (HFMI), project “The Human Radiome Project” (THRP). This work was partially funded by Helmholtz Imaging, a platform of the Helmholtz Information \& Data Science Incubator, by “NUM 2.0“ (FKZ: 01KX2121), by “NUM 3.0” (FKZ: 01KX2524), by the Deutsche Forschungsgemeinschaft (DFG, German Research Foundation) under project number 402688427, and by the Helmholtz AI project “Effective Privacy-Preserving Adaptation of Foundation Models for Medical Tasks” (PAFMIM; ZT-I-PF-5-227). Maximilian Rokuss was supported by the Google PhD Fellowship Program.

\subsubsection{\discintname}
Maximilian Rokuss was supported by the Google PhD Fellowship Program. Other than that the authors have no competing interests to declare that are relevant to the content of this article.
\end{credits}

\bibliographystyle{splncs04}
\bibliography{references}

\end{document}